\documentclass[sigconf,nonacm]{acmart}

\usepackage{booktabs}
\usepackage{tikz}
\usetikzlibrary{positioning,arrows.meta,calc}

\begin{document}

\title{AllPath: Layered Verification for a Deployed, Self-Revising LLM
Trading Agent under Human-Gated and Autonomous Regimes}
\titlenote{Accepted for poster presentation at the 3rd Workshop on LLMs and
Generative AI in Finance (AI4F), ACM ICAIF 2026, Milan (non-archival).}

\author{Tian Zhang}
\email{0tianzhang0@gmail.com}
\affiliation{%
  \institution{Independent Researcher}
  \country{USA}
}

\author{Meng Li}
\email{limengunique0917@gmail.com}
\affiliation{%
  \institution{Independent Researcher}
  \country{USA}
}

\begin{abstract}
LLM trading agents are evaluated almost exclusively in backtests and
autonomous benchmark harnesses. Deployment, where actions are irreversible,
raises a different question: who verifies the agent's actions, and which
failures can each verifier structurally catch? We report on AllPath, a
personal LLM trading agent deployed against a live paper-brokerage account.
The system watches positions through a deterministic rules engine, and each
night it reviews its own day and proposes revisions to its own trading
strategies. Verification is split across three layers. Code verifies what
must not depend on judgment: deterministic gates enforce risk limits and
block any strategy revision that fails an integrity check. The agent
verifies itself within fixed bounds: its nightly review has an iteration
cap and no tools that can act. A human verifies judgment, through approval
screens that show the current state of each proposal. We also compared
two verification policies for the agent's self-revisions on the live
system: two and a half weeks with every self-revision human-gated, then
five weeks with reflection revisions auto-applied behind the same code
gates, while the operator's chat channel stayed human-gated throughout. Each layer caught a different kind of failure. The code
gates rejected no proposal; their role was keeping revisions consistent.
The human rejected two proposals that passed every code check, both over a
disagreement about the market, and later removed 11 of the 19 rules the
agent had added on its own, at a median lag of eight days. The most instructive production incidents, such as strategies silently
left unmonitored, passed every gate; the operator found them by hand, and
the fix each time was an interface change that makes the condition
visible. We
contribute the layered architecture, the two-regime account, and a
production incident taxonomy, none observable in a backtest.
\end{abstract}

\ccsdesc[500]{Computing methodologies~Artificial intelligence}

\keywords{LLM agents, human-in-the-loop, algorithmic trading, AI oversight,
deployed systems}

\maketitle

\section{Introduction}
\label{sec:intro}

LLM agents now trade competently in backtests and simulated
markets~\cite{finmem2024,finagent2024,tradingagents2024,finrobot2024}, and
a recent wave of live-market benchmarks streams them real-time
data~\cite{stockbench2025,livetradebench2025,agentmarketarena2025,aitrader2025}.
Nearly all of this work evaluates the agent as an unconstrained decision
function: its output is the action, and when the run ends, nothing
happened.

Production is less forgiving. In industrial e-commerce search, we found
that an LLM which improves ranking quality offline cannot serve traffic as
it is: latency and stability require distilling it into smaller, more
predictable models before deployment~\cite{shang2025distill,
puthenputhussery2025bert}. Those constraints are about serving. A deployed
trading agent inherits them and adds two of its own. A mis-issued order is
an irreversible fill, not a degraded ranking. And an agent that revises
its own trading rules can compound a bad judgment before anyone looks.

This paper studies that verification problem on a running system.
AllPath\footnote{Code: \url{https://github.com/dukesky/allpath-trading-agent}.
Public daily journal of the deployment (paper account):
\url{https://trading.all-path.com/trading/journal}.} is a personal LLM
trading agent that has run continuously since August 2026 against a live
paper-brokerage account. It watches positions through
a deterministic rules engine, converses with its operator, maintains a
governed memory of stocks, strategies, and distilled lessons
(\S\ref{sec:system}), and each night reviews its own day and proposes
revisions to its own strategies. No single verifier covers all of this.
Whether an order exceeds a risk cap is arithmetic, and belongs in code.
Whether a stop is too tight is judgment, and belongs to a person. The
agent's nightly self-review sits between the two: it is useful only if the
agent cannot act on its own conclusions. \S\ref{sec:stack} assigns every
action in the system to one of these three verifiers.

We also used the deployment to compare two verification policies for the
agent's self-revisions. The system supports both as a configuration
switch: a reflection-proposed revision can wait for human approval, or it
can apply on its own once it passes the code gates. We ran the first
policy for two and a half weeks, during which the operator approved four
proposals and vetoed two. We then switched to the second for five weeks,
with a drawdown circuit breaker as the backstop: 56 revisions applied and
the gates rejected none. The operator's chat channel kept its human gate
under both policies, and through it the operator later removed 11 of the
19 rules the agent had added on its own.

The paper makes three contributions. The first is a layered verification
architecture with production time on it: a decomposition of every action
class in a self-revising trading agent by proposer, verifier, and gate
mechanics (Table~\ref{tab:stack}), with the design rules that make each
layer load-bearing. The second is a two-regime account of self-revision:
operational evidence from the human-gated and autonomous periods,
including the two cases where the human vetoed proposals the gates found
flawless (\S\ref{sec:regimes}). The third is a production incident
taxonomy: twenty incidents from the deployment and its construction, none
observable in a backtest, each generalized to a design rule
(\S\ref{sec:evidence}).

Evaluation throughout is operational: proposal flows, gate activations,
decision latencies, incidents. We know of no prior published account of a
deployed LLM trading agent operated under both human-gated and autonomous
verification; \S\ref{sec:related} gives the basis for that statement and
\S\ref{sec:limitations} its limits.

\section{Related Work}
\label{sec:related}

\paragraph{LLM trading agents.}
A fast-growing line of work equips LLM agents with memory, tools, and
multi-agent structure for trading: layered memory and character design
(TradingGPT, FinMem)~\cite{tradinggpt2023,finmem2024}, multimodal
tool-augmented agents (FinAgent)~\cite{finagent2024}, debate-style
multi-agent frameworks (TradingAgents)~\cite{tradingagents2024}, open agent
platforms (FinRobot)~\cite{finrobot2024}, and simulated markets
(StockAgent)~\cite{stockagent2024}; two surveys map the
field~\cite{finllmsurvey2024,agentictrading2026}. This literature is
backtest- or simulation-centric, and the agent's output is the unconstrained
final action. A newer wave of live-market benchmarks
(StockBench, LiveTradeBench, Agent Market Arena,
AI-Trader)~\cite{stockbench2025,livetradebench2025,agentmarketarena2025,aitrader2025}
streams real-time data but runs agents fully autonomously, several
treating the \emph{absence} of human intervention as a design goal rather
than a variable to study. Our question sits elsewhere: how a deployed
agent's actions get verified.

\paragraph{Self-improvement without stakes.}
Reflexion, Self-Refine, and Voyager established that agents improve by
criticizing and revising their own behavior~\cite{reflexion2023,
selfrefine2023,voyager2023}, and self-evolving-agent surveys now catalogue
the design space~\cite{selfevolvingsurvey2025a,selfevolvingsurvey2025b}.
In these settings a bad self-revision costs a retry; recent work begins to
ask what happens when it costs more~\cite{selfevolvingsafety2026}. Our
reflection loop is a deliberately conservative member of this family: its
sole lever on future behavior is a gated, auditable revision proposal whose
authorization and lifecycle fields are frozen by code (\S\ref{sec:code}).

\paragraph{Agent oversight and governance.}
Scalable-oversight proposals (debate, amplification)~\cite{debate2018,
amplification2018} target verification of systems beyond direct human
evaluation; a complementary risk analysis asks when stacked safety
mechanisms actually fail together~\cite{alignmentrisk2025}. Closer to
practice, runtime-enforcement DSLs (AgentSpec)~\cite{agentspec2025}, guard
agents (GuardAgent)~\cite{guardagent2024}, autonomy-level
taxonomies~\cite{autonomylevels2025}, and interface-mediated oversight
(Magentic-UI; regulable-UI patterns)~\cite{magenticui2025,uigovernance2025}
each propose one layer of the stack we deploy. To our knowledge none are
combined in a single deployed system with consequential actions, and none
report human-gated and autonomous operation on the same live agent.

\paragraph{Trading-system safety.}
Our deterministic layer inherits from pre-LLM market practice:
exchange-level circuit breakers~\cite{subrahmanyam1994} and the SEC's
market-access rule mandating broker-side pre-trade risk
controls~\cite{sec15c3-5}. The contribution is not these mechanisms but
their placement in an agentic loop: caps the agent cannot negotiate with,
a drawdown breaker that demotes autonomy rather than halting trading, and
integrity gates on the agent's self-modification channel, a surface
pre-LLM risk controls never had to consider.

\section{The AllPath System}
\label{sec:system}

\begin{figure*}[t]
\centering
\resizebox{\textwidth}{!}{
\begin{tikzpicture}[
  font=\sffamily\footnotesize,
  every node/.style={align=center},
  comp/.style={draw, rounded corners=3pt, minimum height=10mm, inner sep=4.5pt, line width=0.9pt},
  ext/.style={comp, fill=gray!12, draw=gray!60},
  code/.style={comp, fill=blue!8, draw=blue!55!black},
  gate/.style={comp, fill=blue!16, draw=blue!65!black, line width=1.4pt},
  agent/.style={comp, fill=orange!16, draw=orange!85!black},
  human/.style={comp, fill=green!13, draw=green!50!black, line width=1.4pt},
  store/.style={comp, fill=violet!9, draw=violet!55!black},
  flow/.style={-{Stealth[length=2.8mm]}, line width=1.0pt, draw=black!70},
  autoflow/.style={flow, draw=blue!60!black},
  humanflow/.style={flow, draw=green!45!black},
  lbl/.style={font=\sffamily\scriptsize, inner sep=1pt, fill=white, text=black!80},
]
\node[ext,  minimum width=23mm] (market)   at (0.0,7.0)  {Market data\\\scriptsize live quotes};
\node[code, minimum width=30mm] (sentinel) at (4.4,7.0)  {\textbf{Sentinel}\\\scriptsize deterministic rules engine};
\node[gate, minimum width=27mm] (risk)     at (12.8,7.0) {\textbf{Risk gate}\\\scriptsize per-order / daily / option caps};
\node[ext,  minimum width=23mm] (broker)   at (16.6,7.0) {Broker (paper)\\\scriptsize + shadow ledger};
\node[agent,minimum width=23mm] (chat)     at (0.0,4.9)  {\textbf{Chat agent}\\\scriptsize propose only};
\node[code, minimum width=27mm] (queue)    at (8.4,4.9)  {\textbf{Pending queue}\\\scriptsize single choke point};
\node[human,minimum width=31mm] (operator) at (12.8,4.9) {\textbf{Operator}\\\scriptsize web / Telegram cards\\\scriptsize regenerated diffs $\cdot$ single-use links};
\node[store,minimum width=30mm] (strat)    at (4.4,2.9)  {\textbf{Strategy files}\\\scriptsize versioned $\cdot$ auth + status};
\node[store,minimum width=23mm] (memory)   at (0.0,0.8)  {Layered memory\\\scriptsize guarded writes};
\node[agent,minimum width=32mm] (reflect)  at (4.4,0.8)  {\textbf{Nightly reflection}\\\scriptsize no action tools $\cdot$ iteration cap\\\scriptsize fenced briefing $\cdot$ transcript};
\node[gate, minimum width=31mm] (integ)    at (9.4,0.8)  {\textbf{Revision integrity gates}\\\scriptsize byte-exact base $\cdot$ version$\uparrow$\\\scriptsize auth/status frozen $\cdot$ re-check at apply};
\node[gate, dashed, minimum width=25mm] (breaker) at (15.7,0.8) {\textbf{Drawdown breaker}\\\scriptsize demotes autonomy\\\scriptsize to confirm; manual reset};
\draw[flow]     (market) -- (sentinel);
\draw[autoflow] (sentinel) -- node[lbl, pos=0.5, above=2pt]{\texttt{auto+hard} orders} (risk);
\draw[autoflow] (risk) -- node[lbl, pos=0.35, above=2pt]{execute} (broker);
\draw[flow]     ($(sentinel.south)+(0.6,0)$) |- node[lbl, pos=0.75, above=2pt]{\texttt{confirm} orders} ($(queue.north)+(0,0.4)$) -- ($(queue.north)+(0,0)$);
\draw[flow]     (chat) -- node[lbl, pos=0.68, above=2pt]{propose order / revision} (queue);
\draw[humanflow](queue) -- node[lbl, above=2pt]{review} (operator);
\draw[humanflow](operator) -- node[lbl, right=2pt, pos=0.45]{approve} (risk);
\draw[flow]     (reflect) -- node[lbl, above=2pt, pos=0.42]{propose} (integ);
\draw[flow, dashed] ($(integ.north)+(-0.6,0)$) -- node[lbl, right=3pt, pos=0.5]{human-gated regime:\\\scriptsize enqueue for approval} ($(queue.south)+(0.4,0)$);
\draw[humanflow] (operator.south) -- ++(0,-1.35) -| node[lbl, pos=0.88, right=2pt]{approve} ($(integ.north)+(1.0,0)$);
\draw[autoflow] ($(integ.west)+(0,0.25)$) -- ++(-0.95,0) |- node[lbl, pos=0.25, right=3pt]{apply\\\scriptsize(direct in autonomous\\\scriptsize regime; else via queue)} ($(strat.east)+(0,0.0)$);
\draw[flow]     (strat) -- node[lbl, left=2pt, pos=0.3]{active rules} (sentinel);
\draw[flow, draw=blue!60!black, dashed] (breaker.west) -- (integ.east);
\draw[flow, <->] (memory) -- (reflect);
\begin{scope}[shift={(15.5,4.6)}]
  \node[font=\sffamily\scriptsize\bfseries, anchor=west] at (-0.30,0.62) {Verifier class};
  \node[gate,  minimum width=4.5mm, minimum height=3.4mm] at (0,0) {};
  \node[font=\sffamily\scriptsize, anchor=west] at (0.3,0) {deterministic code};
  \node[agent, minimum width=4.5mm, minimum height=3.4mm] at (0,-0.60) {};
  \node[font=\sffamily\scriptsize, anchor=west] at (0.3,-0.60) {agent, bounded};
  \node[human, minimum width=4.5mm, minimum height=3.4mm] at (0,-1.20) {};
  \node[font=\sffamily\scriptsize, anchor=west] at (0.3,-1.20) {human judgment};
\end{scope}
\end{tikzpicture}}
\caption{AllPath architecture. Every path from a proposal to an irreversible
action crosses at least one verification layer; colors mark the verifier
class (blue: deterministic code gates; orange: agent under enforced bounds;
green: human judgment). Solid blue arrows are fully autonomous paths; green
arrows pass through the operator. The dashed breaker demotes the autonomous
revision path back to human confirmation on excessive drawdown.}
\label{fig:arch}
\end{figure*}
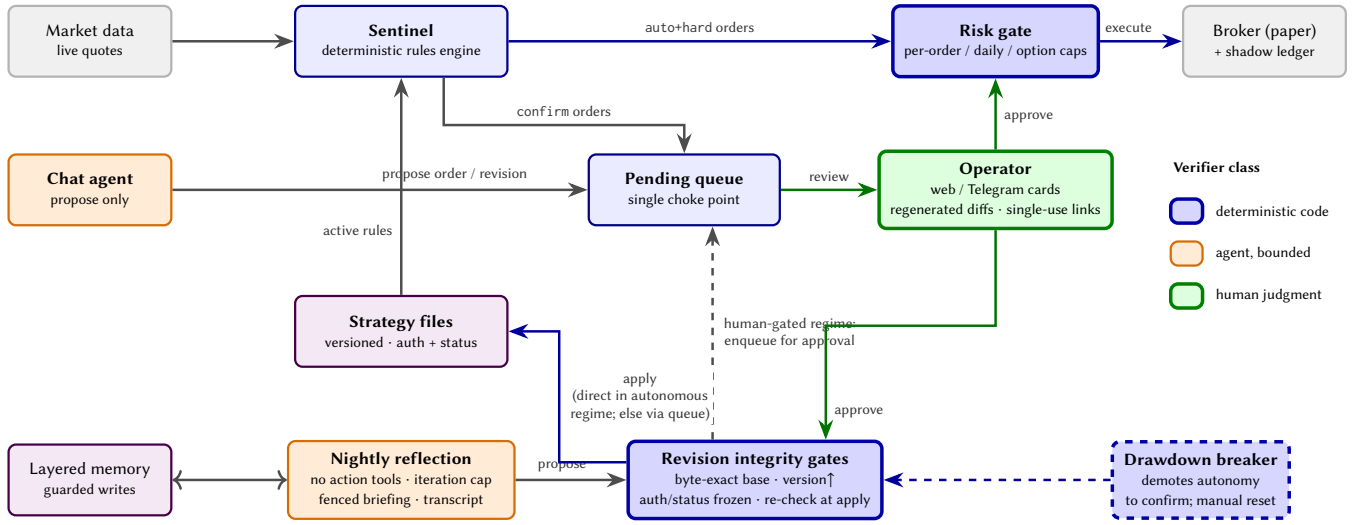

AllPath is a personal LLM trading agent: a deterministic rules engine
watches positions, a chat agent talks to the operator, a nightly
reflection session reviews the day and proposes revisions, and a single
approval queue sits between every proposal and its execution. It is a
single-operator production system rather than a benchmark harness: it
holds positions at live market prices, sends notifications the operator
acts on, and its mistakes cost the operator time. Figure~\ref{fig:arch}
shows the architecture. The verification mechanisms overlaid in the figure
(detailed in \S\ref{sec:stack}) sit at every boundary where an action could
become irreversible.

\subsection{Strategies as versioned files}
A strategy is a human-readable file with four parts: a \emph{thesis} (why
the position exists), \emph{rules} (typed entry, exit, and stop conditions
over prices and indicators), an \emph{authorization} level, and a
\emph{status}. The file is the source of truth. It is versioned on every
change, snapshotted into an append-only history table, and rendered
verbatim to both the human (web UI) and the agent (session context).
Authorization is set per rule consequence: \texttt{notify} tells the human,
\texttt{confirm} queues an order for approval, \texttt{auto} executes within
risk caps, and \texttt{hard} marks a rule as mechanical with no agent
discretion. Status controls monitoring: the sentinel watches only
\texttt{active} strategies. Figure~\ref{fig:strategy} shows the example
file shipped with the repository.

\begin{figure}[t]
\small
\begin{verbatim}
id: example
name: NVDA long-term hold - AI infrastructure play
status: draft
version: 2
authorization: confirm
thesis: 'Bullish on NVDA''s leadership in AI chips
  and data center hardware. 18-24 month horizon.
  [...] Invalidation: AI chip demand deterioration,
  significant market share loss to competitors, or
  datacenter revenue growth <15% for two
  consecutive quarters.'
position:
  ticker: NVDA
  target_weight: '0.2'
  max_weight: '0.25'
rules:
- id: initial-entry
  type: soft
  condition: position_weight == 0 and price < 205
  action: buy $8000
  state: armed
- id: stop-loss
  type: hard
  condition: price < 170
  action: sell all
  state: armed
- id: take-profit-1
  type: soft
  condition: pnl_pct > 40
  action: sell 30%
  state: armed
[...]
\end{verbatim}
\caption{A strategy file (the example shipped with the repository,
abbreviated). Thesis and rules are what a revision may change;
\texttt{authorization} and \texttt{status} are frozen against
self-revision (\S\ref{sec:code}).}
\label{fig:strategy}
\end{figure}

\subsection{Sentinel: the deterministic watcher}
A scheduler daemon evaluates every active strategy's rules against live
quotes on a fixed interval during market hours (between 15 and 60 minutes
over the data window). Rule evaluation is pure
code, with no model in the loop. When a rule fires, the consequence follows
its authorization: \texttt{auto+hard} rules execute directly through the
risk gate, and \texttt{confirm} rules enqueue a pending order with a full
snapshot (trigger price, rule text, intent) for human approval. One-shot
rule state prevents re-triggering; a fired trigger must be re-armed by
hand. The same daemon runs the nightly pipeline of portfolio digest, memory
consolidation, and the reflection session (\S\ref{sec:reflection}), each
step restart-safe behind persisted date markers. Rules may also place
long-only option orders through the same gate; by grammar these are
available to \texttt{auto+hard} rules only, so the option surface is
entirely autonomous.

\subsection{Agent sessions: chat and review}
The \emph{chat agent} is the operator's conversational interface over web
or Telegram. It has read tools for market data, portfolio, journal,
strategy files, and memory search, and a short list of action tools:
propose an order (which enters the same pending queue as sentinel orders),
propose a strategy revision, and guarded memory writes. It cannot place
orders directly, cannot edit strategy files, and cannot approve anything,
including its own proposals. The \emph{review agent} renders pending items
with regenerated context (\S\ref{sec:human}) and records the human's
approve or reject decision; approvals can also arrive through single-use
tokenized links in notifications (\S\ref{sec:code}).

\subsection{Reflection: bounded nightly self-revision}
\label{sec:reflection}
After the close, a separate \emph{reflection session} reviews the day. It
receives a code-generated briefing (fills at execution prices, rule-state
changes, position moves, pending-queue state, full strategy texts, recent
report summaries), investigates with read-only tools under an iteration
cap, and writes a structured report. Its only lever on future behavior is
\texttt{propose\_strategy\_revision}, which may change a strategy's thesis
and rules but never its authorization or status (\S\ref{sec:code}). Under
the \emph{human-gated regime}, every such proposal waits in the queue like
any other. Under the \emph{autonomous regime} (\S\ref{sec:regimes}),
reflection-sourced proposals apply immediately after passing the same
mechanical gates, with a portfolio-drawdown circuit breaker as the
regime-level backstop: if equity drawdown exceeds a configured threshold,
auto-apply demotes to confirm and stays demoted until a human resets it.

\subsection{Memory}
The agent's memory is a set of human-readable markdown files in four
curated layers: an operator profile shared across accounts, per-strategy
notes, per-stock notes, and distilled lessons. Each layer has a size
budget of a few kilobytes, which forces consolidation over accumulation.
The agent has exactly one write path: a guarded tool that makes
entry-level edits, rejects injection-shaped content, and diff-logs every
mutation to the ledger; the operator can read and edit the same files
directly. An append-only observation log records day-to-day events beside
the curated layers. Nightly consolidation distills it into them, and
consolidation markers advance only past cleanly summarized days, so a
crash re-consolidates rather than skips. At the data freeze the paper
account held 141 lessons, 37 stock notes, and 36 strategy notes, all
agent-written under these constraints. Appendix~\ref{app:artifacts} shows
a lesson entry and a reflection report excerpt.

\subsection{Deployment shape}
Production runs on a single consumer machine behind a private network with
no public ports, under a process supervisor that restarts it on failure.
The operator reaches it through a token-authenticated web UI (dashboard,
strategies, pending reviews, reports with replayable reflection
transcripts, journal) and a Telegram channel for notifications, chat, and
approve links. The system runs two accounts through the same agent and the same
pipeline. The \emph{paper} account trades through the broker's
paper-trading API, where the agent may execute on its own. The
\emph{shadow} account is a local ledger mirroring the operator's real
holdings; it never routes an order, and an approved decision is recorded
for the operator to place by hand.

\paragraph{Models.}
Every agent session in the data window ran a Claude-family model through
an API gateway: reflection and memory consolidation on Claude Opus 5, chat
on Claude Sonnet 5, and the sentinel's analysis tier on Claude Haiku 4.5.
Model names are configuration, not code; nothing in the verification
stack depends on which model proposes.

\paragraph{Artifacts.}
The system is open source.\footnote{\url{https://github.com/dukesky/allpath-trading-agent}}
After each trading day's close, the nightly pipeline publishes the paper
account's digest, trades, and reflection write-up to a public journal
page,\footnote{\url{https://trading.all-path.com/trading/journal}} which
has run continuously since Aug 31, 2026 and covers most of the data window
reported in \S\ref{sec:evidence}. The counts in this paper come from the
system's internal ledger, which also holds the operator's strategy files
and is not public; the journal is the day-level record a reader can check
against them.

\section{The Verification Stack}
\label{sec:stack}

The system's design premise is that no single verifier is sufficient for an
agent whose actions are irreversible. We decompose verification by what
kind of failure each verifier is structurally able to catch, and give every
action class an explicit row in that decomposition
(Table~\ref{tab:stack}). Three verifier classes emerge.

\subsection{Code verifies what must never depend on judgment}
\label{sec:code}

Deterministic gates are the only verifiers that cannot be persuaded. We
place them wherever a failure is unambiguous and mechanical.

\begin{itemize}
\item \textbf{Risk gate.} Every order passes a deterministic gate, whether
  it was proposed by the user in chat, fired by a sentinel rule, or (for
  options) selected autonomously. The gate enforces per-order value caps,
  daily trade-count caps, and, for option buys, a premium cap plus a cap on
  total option exposure relative to equity. The gate is code; the agent
  cannot negotiate with it. Closing an option position is never blocked by
  value caps: a close reduces risk, and a gate that can trap the operator
  in a position is itself a hazard.
\item \textbf{Revision integrity gates.} A strategy revision is the
  agent's only path to changing its own trading rules, and it is applied
  only if three checks pass. (i) The current file is \emph{byte-identical}
  to the base the proposal was computed from; any intervening change,
  however small, voids the proposal instead of being silently overwritten.
  (ii) The proposed version strictly increases, which keeps the audit
  trail's ordering truthful. (iii) The proposal changes only
  \texttt{thesis} and \texttt{rules}. \texttt{Authorization} (who may act)
  and \texttt{status} (whether the strategy is live) are frozen: a
  self-revision can tighten a stop, but it can never promote itself to
  autonomous trading or silently de-activate monitoring. A proposal failing
  any check returns to the pending queue. The failure mode is ``nothing
  happened,'' never ``something subtly different happened.''
\item \textbf{Displacement of stale drafts.} If a newer chat-sourced
  revision arrives for a strategy whose older chat draft is still pending,
  the old row is relabeled \texttt{superseded}, its resolution is
  timestamped, and its approval token is invalidated. A stale approve-link
  in a notification can no longer apply an outdated draft, and the audit
  trail records the displacement.
\item \textbf{Idempotency and at-most-once semantics.} The nightly pipeline
  is restart-safe by persisted markers: a report date can be produced only
  once, and memory-consolidation markers advance only past clean summaries.
  Inbound chat messages from the remote channel advance their offset
  \emph{before} processing, so a crash mid-turn loses one visible message
  and never replays an invisible duplicate that could re-propose an order.
\item \textbf{Single-use approval credentials.} Approve-by-link tokens are
  per-review, stored only as hashes, expire in 24\,h, and are consumed on
  first use, including on failed execution. A leaked notification exposes
  at most one already-visible decision, never a session.
\end{itemize}

\subsection{The agent verifies itself, only inside enforced bounds}
\label{sec:self}

The nightly reflection session reviews the day with the agent's full read
toolset and writes a structured report. Its value as self-verification
comes from the bounds.

\begin{itemize}
\item \textbf{No action tools.} The reflection registry contains market,
  portfolio, and journal readers, guarded memory writes, and
  \texttt{propose\_strategy\_revision}. It has no order placement, no
  approval, and no file writes. Tests pin the exclusion by diffing the
  reflection toolset against the action registry.
\item \textbf{Bounded search.} A session gets at most $N$ tool iterations
  (default 12), \emph{including} the one corrective turn allowed when the
  report fails to parse. An earlier draft silently doubled the budget on
  the corrective turn, which review caught; the cap is now a property of
  the session and no longer of the turn.
\item \textbf{Fenced deterministic briefing.} The session opens with a
  code-generated briefing (fills with real execution prices, rule state
  changes, position day-moves, pending-queue state, full strategy texts).
  Every block is wrapped in external-content fences so tool-returned text
  cannot be mistaken for instructions. The agent starts from verified facts
  and spends its budget on investigation.
\item \textbf{Auditable transcript.} Each session is a separate
  conversation whose every tool call is replayable from the Reports page.
  Self-verification that cannot be audited is indistinguishable from
  confabulation. The transcript lets the human spot-check the agent's
  diligence, and it is the basis for the session accounting in
  \S\ref{sec:evidence}.
\end{itemize}

\subsection{The human verifies judgment, if the interface tells the truth}
\label{sec:human}

The human is the only verifier of judgment calls: is this stop too tight,
is this thesis still believed. The system's obligation is to make that
verification real, and several of our incidents (\S\ref{sec:evidence})
were failures of exactly this obligation.

\begin{itemize}
\item \textbf{Regenerated, not stored, diffs.} A pending revision's diff is
  recomputed against the \emph{current} file at render time, with an
  explicit stale-base warning when the file has moved. The human sees what
  approval means now, and not what the agent saw when it proposed. Resolved
  revisions render their frozen historical diff, labeled as such.
\item \textbf{Decision-grade context on approval cards.}
  Table~\ref{tab:card} in Appendix~\ref{app:artifacts} lists every field
  of a card. Each card shows
  the trigger-time price, the live price with day-direction coloring, the
  deviation since trigger, estimated shares for notional orders, and a
  market-state line (``market closed: fills at next open''). These are the
  facts a human needs to catch a no-longer-sensible order at a glance.
\item \textbf{Accurate labels.} Submission time and fill time are distinct
  columns with distinct labels everywhere (journal, dashboards, the agent's
  own system prompt, long-term memory events), after an incident in which a
  submission timestamp presented as a fill time cost the operator a 17-hour
  misunderstanding. ``Fill pending'' is claimed only within the polling
  window; beyond it the system says ``status unconfirmed.''
\item \textbf{Silence made visible.} Strategies whose status excludes them
  from monitoring wear a persistent ``not monitored'' badge on every
  surface. The badge followed the deployment's most instructive incident:
  all four live strategies sat in \texttt{draft} for days with stop-losses
  never armed, while every page rendered them normally. The sentinel
  heartbeat line (``last tick $N$m ago'' or ``monitoring paused: market
  closed'') now states whether anyone is watching, where before the answer
  was left to omission.
\end{itemize}

\subsection{Verifying the builder}
\label{sec:builder}

The stack above was itself constructed under adversarial review. Every
change shipped through an independent reviewer instructed to execute attack
scenarios. Before merge, that review caught one silent-revert Critical (a
stale sibling proposal whose approval would have rolled back an
already-applied stop tightening) and a series of Importants (an
authorization-flip hole, a path-traversal in the revision applier,
resurrection of deleted strategies). We include this as evidence for a
scoping claim: several holes were only findable by \emph{executing} the
failure, and none would have surfaced in a backtest harness.

\begin{table*}[t]
\caption{Every action class in the system, decomposed by proposer, verifier,
and gate mechanics.}
\label{tab:stack}
\small
\begin{tabular}{@{}p{0.21\textwidth}p{0.17\textwidth}p{0.13\textwidth}p{0.41\textwidth}@{}}
\toprule
\textbf{Action} & \textbf{Proposed by} & \textbf{Verified by} & \textbf{Gate mechanics} \\
\midrule
Stock order (\texttt{confirm} auth) & Sentinel rule or chat agent & Code, then human & Risk-gate caps $\to$ pending queue $\to$ human approve (web/link), single-use token \\
Stock order (\texttt{auto} auth, \texttt{hard} rule) & Sentinel rule & Code only & Risk-gate caps; one-shot rule state; drawdown breaker as regime backstop \\
Option order (buy) & Sentinel rule (\texttt{auto+hard} only, by grammar) & Code only & Premium + exposure caps; expiry safety sweep; breaker \\
Option close & Sentinel rule / expiry sweep & Code only & Never value-blocked (risk-reducing); daily count only \\
Strategy revision (human-gated regime) & Reflection or chat agent & Code, then human & Parse + id/version/freeze checks at propose; byte-exact staleness + re-checks at apply; \texttt{superseded} displacement of stale chat drafts; human reads regenerated diff \\
Strategy revision (autonomous regime) & Reflection & Code only & Same mechanical gates, auto-apply; scoped to reflection-sourced rows; breaker backstop \\
Strategy lifecycle (activate/pause) & Human only & Human, with code guardrails & Two-step confirm; atomic write + version snapshot; burned-trigger re-arm warning \\
Memory write & Any agent session & Code (injection guard) & Guarded tool path; fenced external content; layered files \\
Notification / report & System & Code & Never-raise contract; per-channel isolation; failure never blocks the pipeline \\
Trading parameters (caps, models, keys) & Human only & Human & Settings page / env file only; never agent-writable; paper-vs-real flag not web-editable \\
\bottomrule
\end{tabular}
\end{table*}

\section{Self-Revision Under Two Regimes}
\label{sec:regimes}

We ran the self-revision channel under two verification regimes on the
same live deployment, by design: two and a half weeks with a human gate on
every self-proposed change, then five weeks with the mechanical gates
standing alone. This section describes the loop and what each regime looked like in
operation. \S\ref{sec:evidence} reports the numbers.

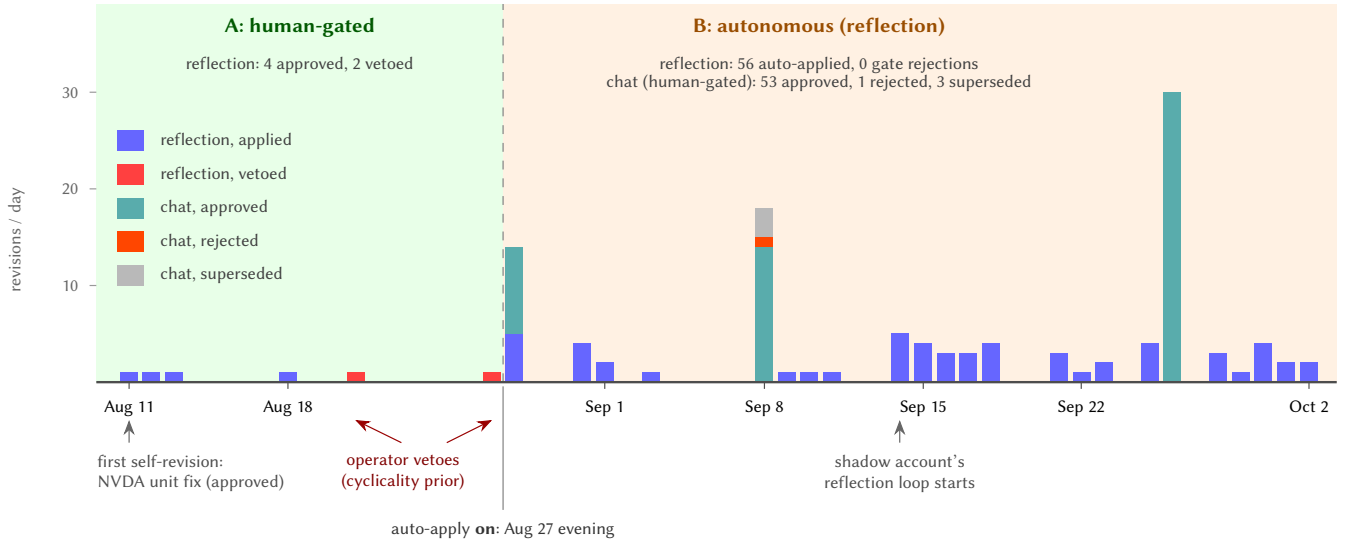
\begin{figure*}[t]
\centering
\resizebox{\textwidth}{!}{
\begin{tikzpicture}[font=\sffamily\scriptsize]
\colorlet{reflApp}{blue!60}
\colorlet{reflRej}{red!75}
\colorlet{chatApp}{teal!65}
\colorlet{chatRej}{red!45!orange}
\colorlet{chatSup}{gray!55}
\fill[green!9]   (-0.25,0) rectangle (4.59,4.5);
\fill[orange!11] (4.59,0) rectangle (14.51,4.5);
\draw[dashed, black!50] (4.59,0) -- (4.59,4.5);
\node[anchor=north, font=\sffamily\footnotesize\bfseries, text=green!35!black] at (2.17,4.45) {A: human-gated};
\node[anchor=north, font=\sffamily\footnotesize\bfseries, text=orange!60!black] at (8.35,4.45) {B: autonomous (reflection)};
\node[anchor=north, text=black!75, align=center] at (2.17,3.98) {reflection: 4 approved, 2 vetoed};
\node[anchor=north, text=black!75, align=center] at (8.35,3.98) {reflection: 56 auto-applied, 0 gate rejections\\[-1pt]chat (human-gated): 53 approved, 1 rejected, 3 superseded};
\fill[reflApp] (0.03,0.000) rectangle (0.24,0.115);
\fill[reflApp] (0.30,0.000) rectangle (0.51,0.115);
\fill[reflApp] (0.57,0.000) rectangle (0.78,0.115);
\fill[reflApp] (1.92,0.000) rectangle (2.13,0.115);
\fill[reflRej] (2.73,0.000) rectangle (2.94,0.115);
\fill[reflRej] (4.35,0.000) rectangle (4.56,0.115);
\fill[reflApp] (4.62,0.000) rectangle (4.83,0.575);
\fill[chatApp] (4.62,0.575) rectangle (4.83,1.610);
\fill[reflApp] (5.43,0.000) rectangle (5.64,0.460);
\fill[reflApp] (5.70,0.000) rectangle (5.91,0.230);
\fill[reflApp] (6.24,0.000) rectangle (6.45,0.115);
\fill[chatApp] (7.59,0.000) rectangle (7.80,1.610);
\fill[chatRej] (7.59,1.610) rectangle (7.80,1.725);
\fill[chatSup] (7.59,1.725) rectangle (7.80,2.070);
\fill[reflApp] (7.86,0.000) rectangle (8.07,0.115);
\fill[reflApp] (8.13,0.000) rectangle (8.34,0.115);
\fill[reflApp] (8.40,0.000) rectangle (8.61,0.115);
\fill[reflApp] (9.21,0.000) rectangle (9.42,0.575);
\fill[reflApp] (9.48,0.000) rectangle (9.69,0.460);
\fill[reflApp] (9.75,0.000) rectangle (9.96,0.345);
\fill[reflApp] (10.02,0.000) rectangle (10.23,0.345);
\fill[reflApp] (10.29,0.000) rectangle (10.50,0.460);
\fill[reflApp] (11.10,0.000) rectangle (11.31,0.345);
\fill[reflApp] (11.37,0.000) rectangle (11.58,0.115);
\fill[reflApp] (11.64,0.000) rectangle (11.85,0.230);
\fill[reflApp] (12.18,0.000) rectangle (12.39,0.460);
\fill[chatApp] (12.45,0.000) rectangle (12.66,3.450);
\fill[reflApp] (12.99,0.000) rectangle (13.20,0.345);
\fill[reflApp] (13.26,0.000) rectangle (13.47,0.115);
\fill[reflApp] (13.53,0.000) rectangle (13.74,0.460);
\fill[reflApp] (13.80,0.000) rectangle (14.01,0.230);
\fill[reflApp] (14.07,0.000) rectangle (14.28,0.230);
\draw[black!70, line width=0.8pt] (-0.25,0) -- (14.51,0);
\draw[black!70] (0.14,0) -- (0.14,-0.08);
\node[anchor=north] at (0.14,-0.1) {Aug 11};
\draw[black!70] (2.03,0) -- (2.03,-0.08);
\node[anchor=north] at (2.03,-0.1) {Aug 18};
\draw[black!70] (5.80,0) -- (5.80,-0.08);
\node[anchor=north] at (5.80,-0.1) {Sep 1};
\draw[black!70] (7.70,0) -- (7.70,-0.08);
\node[anchor=north] at (7.70,-0.1) {Sep 8};
\draw[black!70] (9.59,0) -- (9.59,-0.08);
\node[anchor=north] at (9.59,-0.1) {Sep 15};
\draw[black!70] (11.47,0) -- (11.47,-0.08);
\node[anchor=north] at (11.47,-0.1) {Sep 22};
\draw[black!70] (14.18,0) -- (14.18,-0.08);
\node[anchor=north] at (14.18,-0.1) {Oct 2};
\draw[black!40] (-0.25,1.15) -- (-0.33,1.15);
\node[anchor=east, text=black!60] at (-0.35,1.15) {10};
\draw[black!40] (-0.25,2.30) -- (-0.33,2.30);
\node[anchor=east, text=black!60] at (-0.35,2.30) {20};
\draw[black!40] (-0.25,3.45) -- (-0.33,3.45);
\node[anchor=east, text=black!60] at (-0.35,3.45) {30};
\node[anchor=center, rotate=90, text=black!60] at (-1.20,1.7) {revisions / day};
\draw[-{Stealth[length=1.8mm]}, black!60] (0.14,-0.72) -- (0.14,-0.45);
\node[anchor=north west, align=left, text=black!70] at (-0.35,-0.75) {first self-revision:\\NVDA unit fix (approved)};
\draw[-{Stealth[length=1.8mm]}, red!60!black] (3.38,-0.72) -- (2.83,-0.45);
\draw[-{Stealth[length=1.8mm]}, red!60!black] (3.92,-0.72) -- (4.46,-0.45);
\node[anchor=north, align=center, text=red!50!black] at (3.40,-0.75) {operator vetoes\\(cyclicality prior)};
\draw[-{Stealth[length=1.8mm]}, black!60] (9.31,-0.72) -- (9.31,-0.45);
\node[anchor=north, align=center, text=black!70] at (9.31,-0.75) {shadow account's\\reflection loop starts};
\node[anchor=north, align=center, text=black!80] at (4.59,-1.55) {auto-apply \textbf{on}: Aug 27 evening};
\draw[black!50] (4.59,-1.52) -- (4.59,-0.05);
\fill[reflApp] (0.00,2.75) rectangle (0.32,2.99);
\node[anchor=west, text=black!75] at (0.40,2.87) {reflection, applied};
\fill[reflRej] (0.00,2.35) rectangle (0.32,2.59);
\node[anchor=west, text=black!75] at (0.40,2.47) {reflection, vetoed};
\fill[chatApp] (0.00,1.95) rectangle (0.32,2.19);
\node[anchor=west, text=black!75] at (0.40,2.07) {chat, approved};
\fill[chatRej] (0.00,1.55) rectangle (0.32,1.79);
\node[anchor=west, text=black!75] at (0.40,1.67) {chat, rejected};
\fill[chatSup] (0.00,1.15) rectangle (0.32,1.39);
\node[anchor=west, text=black!75] at (0.40,1.27) {chat, superseded};
\end{tikzpicture}}
\caption{Every strategy revision in the data window (Aug 11 to Oct 2, both
accounts), by day, source, and outcome. Regime B auto-applies
\emph{reflection}-sourced proposals only; the chat channel stays
human-gated throughout, which gives a within-regime contrast.}
\label{fig:timeline}
\end{figure*}

\subsection{The revision loop}

A revision proposal, from the nightly reflection session or from the
operator's chat agent, is validated at propose time: it must parse, target
an existing strategy, strictly increase the version, and leave
authorization and status untouched. It then enters the pending queue with
a snapshot of the base file it was computed from. At apply time the
integrity gates run again, now against the live file: byte-exact
staleness, version monotonicity, field freeze (\S\ref{sec:code}). The two
regimes differ only in who pulls the trigger between those two
checkpoints.

\subsection{Regime A: human-gated (Aug 11--27)}

Every proposal waited for the operator. Six reflection-sourced proposals
arrived in this window. Four were approved and two rejected, and the
mechanical gates raised no objection to either rejected proposal. Both
shared a shape: after an upward price move had left pullback-only entry
ladders unreachable, the agent proposed adding a near-market buy tranche
(in one case also raising position-weight caps), with a written rationale
framing the add as a support retest and not momentum chasing. The operator
declined both for a reason that appears nowhere in the proposals: a prior
belief that semiconductors are cyclical, that the ongoing rally was a
swing within a cycle and not a durable re-rating, and that the correct
response to an unreachable ladder was to wait for the cycle to revisit it.
\footnote{Operator's recollection, recorded 2026-09-12; the rejections
themselves are timestamped in the review ledger.} The disagreement locates
the human gate's distinct contribution. The agent's argument was
internally coherent and mechanically compliant, and the veto came from a
world-model the agent did not hold. No gate below the human layer is
structured to adjudicate between two self-consistent theses.

The window's signature case is the deployment's first self-revision. The
reflection session, auditing a newly transcribed NVDA strategy, found that
its scale-in ladder expressed position-weight caps in percent
(\texttt{position\_weight < 12}) while the strategy's target weight was a
decimal (\texttt{0.2}). The caps were vacuously true and the ladder's
risk limit silently void. The proposed fix normalized the units
(\texttt{< 0.16}, \texttt{< 0.2}) and documented the convention in the
thesis. It also removed the momentum-add rule, reasoning from that day's
fill: a breakout add had lifted average cost 2.4\% while leaving the
pullback ladder unreachable. The operator approved it as proposed. This is
the loop working as designed: a consequential bug, of a kind a backtest
cannot see because the file had never been executed against, found by
bounded self-review and fixed through a human gate.

\subsection{Regime B: autonomous (Aug 27 evening--Oct 2)}

One configuration flag scoped the change. From the evening of Aug 27,
reflection-sourced proposals that pass the mechanical gates apply
immediately, with no human in the loop. The operator's chat channel kept
its human gate throughout, so the two verification paths ran side by side
on the same accounts under the same market. From Sep 14 the shadow account
ran its own nightly reflection under the same flag.

In this regime the reflection channel auto-applied 56 revisions, 32 on the
paper account and 24 on the shadow account, each within about two minutes
of being proposed. The integrity gates rejected none. The human-gated chat
channel processed 57 proposals over the same weeks: 53 approved, 1
rejected, and 3 displaced as \texttt{superseded}. The displacement gate
fired on Sep 8, when three stale chat drafts were relabeled instead of
silently replaced (\S\ref{sec:code}).

\subsection{The regime-level backstop}

Autonomy was conditioned on a drawdown circuit breaker. If account equity
falls more than a configured fraction from its tracked peak, auto-apply
demotes to \texttt{confirm}: the queue and gates keep operating, but a
human is back in the loop. The tripped state persists until a human resets
it, and there is no automatic re-arm. The breaker stayed armed for the
whole regime and never tripped, so the backstop is part of the design but
untested by events.

\subsection{Reading the comparison}

The regimes are sequential and the market moved between them. Two
contrasts are available: regime A against regime B on the reflection
channel, and, inside regime B, the autonomous reflection channel against
the human-gated chat channel. \S\ref{sec:evidence} quantifies both, and
\S\ref{sec:limitations} states what they cannot show.

\section{Operational Evidence}
\label{sec:evidence}

\subsection{The revision pipeline, quantified}

Figure~\ref{fig:timeline} shows every strategy revision in the data window
by day, source, and outcome. Through Oct~2 the system processed 119
revision proposals: regime~A's six reflection proposals (4 approved, 2
vetoed), then regime~B's 56 auto-applied reflection revisions alongside 57
human-gated chat proposals (53 approved, 1 rejected, 3 displaced as
\texttt{superseded}).

Two operational facts frame the regime comparison. First, human
verification has latency. The six regime~A decisions took a median of
8.8 hours from proposal to resolution, with a range of 11 minutes to 80.7
hours; the maximum spanned a weekend. A nightly self-revision loop gated by a human
is in practice a next-morning loop, which is acceptable for thesis-level
edits and was part of the motivation for testing the autonomous regime. In
regime~B the same class of proposal applied within about two minutes.
Second, every veto in the window came from a human: two on reflection
proposals in regime~A and one on a chat proposal in regime~B. The
mechanical gates rejected nothing that reached them. Their observed
contribution was integrity, in displacing three stale drafts and holding
authorization frozen, and not filtering. The layers caught disjoint
failure classes, which is the design premise.

The order channel is thinner. Twelve stock orders passed through the
approval queue, five in Aug 10--18 and seven in Sep 14--28; the operator
approved eleven and rejected one. All other order flow came from
\texttt{auto+hard} rules and met only the code layer. The regime
comparison in this paper therefore rests on the revision channel.

\subsection{What the autonomous revisions changed, and what the human later undid}
\label{sec:undo}

The 56 auto-applied revisions touched 20 strategies. Every one rewrote the
strategy's thesis; 49 also edited existing rules (67 condition edits, 19
action edits, 12 state changes), 16 added rules (19 in total), and 6
removed rules (11 in total). Among edits to sell-side price thresholds the
direction was mixed: four moved a stop closer to the market and five moved
one further away. So the autonomous channel was not a one-way ratchet in
either direction; it re-anchored levels as prices moved.

The more telling number is what happened afterwards. Of the 19 rules the
agent added on its own authority, 11 were later removed by the operator
through the human-gated chat channel, at a median lag of 8 days (range 1
to 17). The largest single case: on Aug 31 the reflection session added a
\texttt{salvage-options} exit rule to four strategies at once; on Sep 8
the operator removed it from all four. One further agent-added rule pair
was removed by a later reflection session itself. No modified condition
was ever reverted to its previous value.

We read this as oversight moving in time rather than disappearing. With
the gate on the reflection channel removed, the human did not stop
reviewing the agent's rule changes; the review happened after the fact,
roughly a week later, through a different channel. The integrity gates
could not have intervened, since every one of these rules was well formed.
A removal is not proof the rule was wrong (the operator may restructure a
strategy for other reasons), so we report the counts and the lag and not a
verdict.

\subsection{Cases}

Table~\ref{tab:cases} grounds the abstractions in the deployment's actual
decisions: for each representative case, who proposed, who verified, and
what happened.

\begin{table*}[t]
\caption{Representative cases from the deployment window: every consequential
action, by proposer, verifier, and outcome. $\checkmark$ = approved/executed.}
\label{tab:cases}
\small
\begin{tabular}{@{}p{0.13\textwidth}p{0.27\textwidth}p{0.11\textwidth}p{0.13\textwidth}p{0.27\textwidth}@{}}
\toprule
\textbf{When} & \textbf{Action} & \textbf{Proposer} & \textbf{Verifier} & \textbf{Outcome} \\
\midrule
Aug 10--Sep 28 & 12 stock orders under \texttt{confirm} authorization & Sentinel / chat & Code $\to$ human & 11 $\checkmark$ executed or recorded via the risk gate; 1 rejected \\
Aug 11 & NVDA revision v1$\to$v2: unit-inconsistency fix (\texttt{<12} vs.\ \texttt{0.2}), momentum-add rule removed & Reflection & Code $\to$ human & $\checkmark$ applied; inert position-weight caps made effective \\
Aug 21, 27 & Near-market add tranches after run-ups (TSM \$440 step; NVDA gap-retest \texttt{<\$215} with cap raises) & Reflection & Code $\to$ human & \textbf{Vetoed} on the operator's cyclicality prior; mechanically valid \\
Aug 28--Oct 2 & 56 strategy revisions across both accounts & Reflection & Code only (autonomous) & All auto-applied within about two minutes; 0 integrity-gate rejections \\
Aug 28--Oct 2 & 57 strategy revisions & Chat & Code $\to$ human & 53 $\checkmark$; 1 rejected; 3 superseded \\
Sep 8 & 3 stale chat drafts displaced by newer proposals & Chat & Code (displacement) & Relabeled \texttt{superseded}; approval tokens invalidated \\
Window-wide & Stock stop-loss exits under \texttt{auto+hard} rules & Sentinel & Code only & Executed at market; one-shot state prevented re-fires \\
Regime B & Option close rules and expiry sweep during a 2.5\,h environment outage (\S\ref{sec:incidents}) & Sentinel & Code only & 18 logged error attempts, then closes succeeded; stock leg had exited 2.5\,h earlier \\
\bottomrule
\end{tabular}
\end{table*}

\subsection{Incident taxonomy}
\label{sec:incidents}

We catalogued 20 defects and incidents from the deployment period and its
construction. Six surfaced \emph{in production operation}; thirteen were
caught \emph{pre-merge} by the adversarial review under which the system
was built (\S\ref{sec:builder}); one, the authorization freeze, is a design
decision recorded alongside them because it closed a hole the review had
named. Table~\ref{tab:incidents} lists the most
instructive, each generalized to a design rule. None of the production six
would appear in a backtest: they live in the gap between ``the logic is
right'' and ``the deployed system, its broker, its dependencies, and its
operator interfaces are telling each other the truth.''

Three deserve a sentence of narrative. In the \emph{draft-never-monitored}
trap, all four live strategies sat in \texttt{draft} status for days with
stop-losses unarmed while every page rendered them normally. It is the
incident that motivated the interface rules of \S\ref{sec:human}.
The \emph{submitted-vs-filled drift} cost the operator a 17-hour
misunderstanding: a Monday-16:27 order submission displayed as if it were
a fill; the position actually filled Tuesday 09:34. And the
\emph{id-less success payload}: during the autonomous window, a broker
response with HTTP 200 but no order id was journaled as ``submitted,''
burning five one-shot rules against trades that never existed. The
operator caught it by reading the journal; no gate did. The fix treats a
payload without an order id as a hard error.

\begin{table*}[t]
\caption{Incident taxonomy (selection from 20 catalogued). ``Adversarial
review'' = caught pre-merge during construction (\S\ref{sec:builder});
``production'' = surfaced live during the deployment window.}
\label{tab:incidents}
\small
\begin{tabular}{@{}p{0.36\textwidth}p{0.17\textwidth}p{0.41\textwidth}@{}}
\toprule
\textbf{Incident} & \textbf{Caught by} & \textbf{Design rule derived} \\
\midrule
Draft strategies silently excluded from monitoring; no surface said so & Operator, production & Any state that disables monitoring wears a persistent badge on every surface \\
Submission timestamp presented as fill time (17\,h misread) & Operator, production & Submitted and filled are distinct events with distinct labels, everywhere \\
Id-less HTTP-200 broker payload journaled as ``submitted''; 5 one-shot rules burned & Operator, production (autonomous window) & Validate response content, not shape, before recording success \\
External CLI dependency vanished; 2.5\,h option-close retry storm while the stock leg had already exited & Code (contained), production & Preflight external dependencies; alert distinctly from routine order errors \\
Quote feed returned non-finite \texttt{previous\_close} & Code gate, production & Strict ingestion validation; loud isolated failure beats silent \texttt{NaN} comparison \\
Unparseable sentinel analysis & Code default, production & Fail closed to human review; never default-execute on unparseable agent output \\
Stale sibling proposal would silently revert an applied stop-tightening & Adversarial review & Byte-exact re-check of live state at approval time, not at proposal time \\
TOCTOU race allowing double execution on order claim & Adversarial review & Atomic claim at every irreversible-action gate \\
Corrective turn silently doubled the reflection iteration budget & Adversarial review & A cap is a session property with a single enforcement point covering retries \\
NUL byte in a chat message caused a persistent 500 & Adversarial review & Sanitize model output and user input against internal escaping schemes \\
\bottomrule
\end{tabular}
\end{table*}

\subsection{Reflection reliability and accounting}

All 52 nightly reflection sessions in the window, 37 on the paper account
and 15 on the shadow account, completed and produced a parseable report.
Stored transcripts run 23 to 44 messages per session (median 35) and are
replayable from the Reports page. One accounting gap remains: per-session
token metering was instrumented for the chat and memory tiers but not for
reflection, so we report session sizes and outcomes instead of dollar
costs.

\section{Limitations}
\label{sec:limitations}

\textbf{One system, one operator, who is also an author.} This is a
single-case study, and the human verifier whose judgment we analyze is the
paper's first author. The rejection rationale in \S\ref{sec:regimes} is
self-reported (contemporaneously ledgered decisions, retrospectively
explained). Every count, diff, and timestamp comes from the system's
append-only ledger, but selection and interpretation are ours.

\textbf{Paper trading.} Fills are real-market but capital is not at risk;
the operator's stakes are attention and pride, not money. A human gate
under financial stress may behave differently in both directions (more
vigilant, or more anxious and overriding). The code gates are indifferent
to this distinction; the human-layer findings should be read as a lower
bound on the phenomenon, not a measurement of it. We do not report the account's return. No claim in
this paper depends on its sign, and a 7.5-week window on medium-horizon
strategies could not statistically support a performance claim in either
direction. This is why the evaluation is operational.

\textbf{A natural experiment, not an experiment.} The regimes are
sequential; August's rally and September's tape differ; proposal mix
shifted with the market. The within-B contrast between the chat and
reflection channels shares one market, which bounds the confound for that
comparison only; the A-versus-B comparison carries it in full. We accordingly make
existence claims (``the human gate vetoed proposals no mechanical gate
could,'' ``these incidents occurred and were caught here'') and report
descriptive statistics; we do not claim one regime is measurably safer.

\textbf{Scale and scope.} A 7.5-week window, 119 revisions, one broker,
one model family, one asset mix (US equities plus long-only options). The
breaker never tripped, so the regime-level backstop is untested by events.
Only twelve orders passed through the human queue
(\S\ref{sec:evidence}), so the regime comparison rests on the revision
channel.

\textbf{What we believe generalizes} is the shape, not the thresholds: the
decomposition of verification by failure class (Table~\ref{tab:stack}),
the freeze on self-expanding autonomy, accurate interfaces as a
verification requirement, and the incident classes of
Table~\ref{tab:incidents}. Each is a property of deployed LLM agents with
irreversible actions, and none is specific to this codebase.

\section{Conclusion}
\label{sec:conclusion}

We asked who verifies a trading agent, and answered with a deployed
system's decomposition. Code verifies what must never depend on judgment.
The agent verifies itself only inside enforced bounds. The human verifies
judgment, through interfaces that must show the true state of the system.
Running the same live agent human-gated and then autonomous showed the
layers catching disjoint failures: the code gates kept revisions consistent
and rejected nothing, and the human twice vetoed a proposal over a view of
the market. The production incidents that taught us the most passed every gate and
were found by the operator by hand; each fix was an interface that makes
the condition visible. For agents whose
actions are irreversible, verification is not a layer you add but a
decomposition you design, and the human's place in it is judgment, which
nothing below can absorb.

\bibliographystyle{ACM-Reference-Format}
\bibliography{references}


\begin{thebibliography}{30}


\ifx \showCODEN    \undefined \def \showCODEN     #1{\unskip}     \fi
\ifx \showDOI      \undefined \def \showDOI       #1{#1}\fi
\ifx \showISBNx    \undefined \def \showISBNx     #1{\unskip}     \fi
\ifx \showISBNxiii \undefined \def \showISBNxiii  #1{\unskip}     \fi
\ifx \showISSN     \undefined \def \showISSN      #1{\unskip}     \fi
\ifx \showLCCN     \undefined \def \showLCCN      #1{\unskip}     \fi
\ifx \shownote     \undefined \def \shownote      #1{#1}          \fi
\ifx \showarticletitle \undefined \def \showarticletitle #1{#1}   \fi
\ifx \showURL      \undefined \def \showURL       {\relax}        \fi
\providecommand\bibfield[2]{#2}
\providecommand\bibinfo[2]{#2}
\providecommand\natexlab[1]{#1}
\providecommand\showeprint[2][]{arXiv:#2}

\bibitem[Chen et~al\mbox{.}(2025)]%
        {stockbench2025}
\bibfield{author}{\bibinfo{person}{Yanxu Chen}, \bibinfo{person}{Zijun Yao},
  \bibinfo{person}{Yantao Liu}, \bibinfo{person}{Amy Xin}, \bibinfo{person}{Jin
  Ye}, \bibinfo{person}{Jianing Yu}, \bibinfo{person}{Lei Hou}, {and}
  \bibinfo{person}{Juanzi Li}.} \bibinfo{year}{2025}\natexlab{}.
\newblock \showarticletitle{StockBench: Can LLM Agents Trade Stocks Profitably
  In Real-world Markets?}
\newblock \bibinfo{journal}{\emph{arXiv preprint arXiv:2510.02209}}
  (\bibinfo{year}{2025}).
\newblock


\bibitem[Christiano et~al\mbox{.}(2018)]%
        {amplification2018}
\bibfield{author}{\bibinfo{person}{Paul Christiano}, \bibinfo{person}{Buck
  Shlegeris}, {and} \bibinfo{person}{Dario Amodei}.}
  \bibinfo{year}{2018}\natexlab{}.
\newblock \showarticletitle{Supervising Strong Learners by Amplifying Weak
  Experts}.
\newblock \bibinfo{journal}{\emph{arXiv preprint arXiv:1810.08575}}
  (\bibinfo{year}{2018}).
\newblock


\bibitem[Ding et~al\mbox{.}(2024)]%
        {finllmsurvey2024}
\bibfield{author}{\bibinfo{person}{Han Ding}, \bibinfo{person}{Yinheng Li},
  \bibinfo{person}{Junhao Wang}, \bibinfo{person}{Hang Chen},
  \bibinfo{person}{Doudou Guo}, {and} \bibinfo{person}{Yunbai Zhang}.}
  \bibinfo{year}{2024}\natexlab{}.
\newblock \showarticletitle{Large Language Model Agent in Financial Trading: A
  Survey}.
\newblock \bibinfo{journal}{\emph{arXiv preprint arXiv:2408.06361}}
  (\bibinfo{year}{2024}).
\newblock


\bibitem[Dung and Mai(2025)]%
        {alignmentrisk2025}
\bibfield{author}{\bibinfo{person}{Leonard Dung} {and} \bibinfo{person}{Florian
  Mai}.} \bibinfo{year}{2025}\natexlab{}.
\newblock \showarticletitle{AI Alignment Strategies from a Risk Perspective:
  Independent Safety Mechanisms or Shared Failures?}
\newblock \bibinfo{journal}{\emph{arXiv preprint arXiv:2510.11235}}
  (\bibinfo{year}{2025}).
\newblock


\bibitem[Fan et~al\mbox{.}(2025)]%
        {aitrader2025}
\bibfield{author}{\bibinfo{person}{Tianyu Fan}, \bibinfo{person}{Yuhao Yang},
  \bibinfo{person}{Yangqin Jiang}, \bibinfo{person}{Yifei Zhang},
  \bibinfo{person}{Yuxuan Chen}, {and} \bibinfo{person}{Chao Huang}.}
  \bibinfo{year}{2025}\natexlab{}.
\newblock \showarticletitle{AI-Trader: Benchmarking Autonomous Agents in
  Real-Time Financial Markets}.
\newblock \bibinfo{journal}{\emph{arXiv preprint arXiv:2512.10971}}
  (\bibinfo{year}{2025}).
\newblock


\bibitem[Fang et~al\mbox{.}(2025)]%
        {selfevolvingsurvey2025a}
\bibfield{author}{\bibinfo{person}{Jinyuan Fang}, \bibinfo{person}{Yanwen
  Peng}, \bibinfo{person}{Xi Zhang}, \bibinfo{person}{Yingxu Wang},
  \bibinfo{person}{Xinhao Yi}, \bibinfo{person}{Guibin Zhang},
  \bibinfo{person}{Yi Xu}, \bibinfo{person}{Bin Wu}, \bibinfo{person}{Siwei
  Liu}, \bibinfo{person}{Zihao Li}, \bibinfo{person}{Zhaochun Ren},
  \bibinfo{person}{Nikos Aletras}, \bibinfo{person}{Xi Wang},
  \bibinfo{person}{Han Zhou}, {and} \bibinfo{person}{Zaiqiao Meng}.}
  \bibinfo{year}{2025}\natexlab{}.
\newblock \showarticletitle{A Comprehensive Survey of Self-Evolving AI Agents:
  A New Paradigm Bridging Foundation Models and Lifelong Agentic Systems}.
\newblock \bibinfo{journal}{\emph{arXiv preprint arXiv:2508.07407}}
  (\bibinfo{year}{2025}).
\newblock


\bibitem[Feng et~al\mbox{.}(2025a)]%
        {uigovernance2025}
\bibfield{author}{\bibinfo{person}{K.~J.~Kevin Feng}, \bibinfo{person}{Tae~Soo
  Kim}, \bibinfo{person}{Rock~Yuren Pang}, \bibinfo{person}{Faria Huq},
  \bibinfo{person}{Tal August}, {and} \bibinfo{person}{Amy~X. Zhang}.}
  \bibinfo{year}{2025}\natexlab{a}.
\newblock \showarticletitle{On the Regulatory Potential of User Interfaces for
  AI Agent Governance}.
\newblock \bibinfo{journal}{\emph{arXiv preprint arXiv:2512.00742}}
  (\bibinfo{year}{2025}).
\newblock
\newblock
\shownote{RegML Workshop at NeurIPS 2025 (oral)}.


\bibitem[Feng et~al\mbox{.}(2025b)]%
        {autonomylevels2025}
\bibfield{author}{\bibinfo{person}{K.~J.~Kevin Feng}, \bibinfo{person}{David~W.
  McDonald}, {and} \bibinfo{person}{Amy~X. Zhang}.}
  \bibinfo{year}{2025}\natexlab{b}.
\newblock \showarticletitle{Levels of Autonomy for AI Agents}.
\newblock \bibinfo{journal}{\emph{arXiv preprint arXiv:2506.12469}}
  (\bibinfo{year}{2025}).
\newblock


\bibitem[Gao et~al\mbox{.}(2025)]%
        {selfevolvingsurvey2025b}
\bibfield{author}{\bibinfo{person}{Huan-ang Gao}, \bibinfo{person}{Jiayi Geng},
  \bibinfo{person}{Wenyue Hua}, \bibinfo{person}{Mengkang Hu},
  \bibinfo{person}{Xinzhe Juan}, \bibinfo{person}{Hongzhang Liu},
  \bibinfo{person}{Shilong Liu}, \bibinfo{person}{Jiahao Qiu},
  \bibinfo{person}{Xuan Qi}, \bibinfo{person}{Yiran Wu},
  \bibinfo{person}{Hongru Wang}, \bibinfo{person}{Han Xiao},
  \bibinfo{person}{Yuhang Zhou}, \bibinfo{person}{Shaokun Zhang},
  \bibinfo{person}{Jiayi Zhang}, \bibinfo{person}{Jinyu Xiang},
  \bibinfo{person}{Yixiong Fang}, \bibinfo{person}{Qiwen Zhao},
  \bibinfo{person}{Dongrui Liu}, \bibinfo{person}{Qihan Ren},
  \bibinfo{person}{Cheng Qian}, \bibinfo{person}{Zhenhailong Wang},
  \bibinfo{person}{Minda Hu}, \bibinfo{person}{Huazheng Wang},
  \bibinfo{person}{Qingyun Wu}, \bibinfo{person}{Heng Ji}, {and}
  \bibinfo{person}{Mengdi Wang}.} \bibinfo{year}{2025}\natexlab{}.
\newblock \showarticletitle{A Survey of Self-Evolving Agents: What, When, How,
  and Where to Evolve on the Path to Artificial Super Intelligence}.
\newblock \bibinfo{journal}{\emph{arXiv preprint arXiv:2507.21046}}
  (\bibinfo{year}{2025}).
\newblock


\bibitem[Irving et~al\mbox{.}(2018)]%
        {debate2018}
\bibfield{author}{\bibinfo{person}{Geoffrey Irving}, \bibinfo{person}{Paul
  Christiano}, {and} \bibinfo{person}{Dario Amodei}.}
  \bibinfo{year}{2018}\natexlab{}.
\newblock \showarticletitle{AI Safety via Debate}.
\newblock \bibinfo{journal}{\emph{arXiv preprint arXiv:1805.00899}}
  (\bibinfo{year}{2018}).
\newblock


\bibitem[Li et~al\mbox{.}(2023)]%
        {tradinggpt2023}
\bibfield{author}{\bibinfo{person}{Yang Li}, \bibinfo{person}{Yangyang Yu},
  \bibinfo{person}{Haohang Li}, \bibinfo{person}{Zhi Chen}, {and}
  \bibinfo{person}{Khaldoun Khashanah}.} \bibinfo{year}{2023}\natexlab{}.
\newblock \showarticletitle{TradingGPT: Multi-Agent System with Layered Memory
  and Distinct Characters for Enhanced Financial Trading Performance}.
\newblock \bibinfo{journal}{\emph{arXiv preprint arXiv:2309.03736}}
  (\bibinfo{year}{2023}).
\newblock


\bibitem[Lin et~al\mbox{.}(2026)]%
        {selfevolvingsafety2026}
\bibfield{author}{\bibinfo{person}{Ruixiao Lin}, \bibinfo{person}{Xinhao Deng},
  \bibinfo{person}{Qingming Li}, \bibinfo{person}{Jianan Ma},
  \bibinfo{person}{Yunhao Feng}, \bibinfo{person}{Yuqi Qing},
  \bibinfo{person}{Zhenyuan Li}, \bibinfo{person}{Yechao Zhang},
  \bibinfo{person}{Shiwen Cui}, \bibinfo{person}{Changhua Meng},
  \bibinfo{person}{Tianwei Zhang}, \bibinfo{person}{Xingjun Ma},
  \bibinfo{person}{Qi Li}, \bibinfo{person}{Ke Xu}, {and}
  \bibinfo{person}{Shouling Ji}.} \bibinfo{year}{2026}\natexlab{}.
\newblock \showarticletitle{Safety in Self-Evolving LLM Agent Systems: Threats,
  Amplification, and Case Studies}.
\newblock \bibinfo{journal}{\emph{arXiv preprint arXiv:2606.23075}}
  (\bibinfo{year}{2026}).
\newblock


\bibitem[Madaan et~al\mbox{.}(2023)]%
        {selfrefine2023}
\bibfield{author}{\bibinfo{person}{Aman Madaan}, \bibinfo{person}{Niket
  Tandon}, \bibinfo{person}{Prakhar Gupta}, \bibinfo{person}{Skyler Hallinan},
  \bibinfo{person}{Luyu Gao}, \bibinfo{person}{Sarah Wiegreffe},
  \bibinfo{person}{Uri Alon}, \bibinfo{person}{Nouha Dziri},
  \bibinfo{person}{Shrimai Prabhumoye}, \bibinfo{person}{Yiming Yang},
  \bibinfo{person}{Shashank Gupta}, \bibinfo{person}{Bodhisattwa~Prasad
  Majumder}, \bibinfo{person}{Katherine Hermann}, \bibinfo{person}{Sean
  Welleck}, \bibinfo{person}{Amir Yazdanbakhsh}, {and} \bibinfo{person}{Peter
  Clark}.} \bibinfo{year}{2023}\natexlab{}.
\newblock \showarticletitle{Self-Refine: Iterative Refinement with
  Self-Feedback}.
\newblock \bibinfo{journal}{\emph{arXiv preprint arXiv:2303.17651}}
  (\bibinfo{year}{2023}).
\newblock


\bibitem[Mozannar et~al\mbox{.}(2025)]%
        {magenticui2025}
\bibfield{author}{\bibinfo{person}{Hussein Mozannar}, \bibinfo{person}{Gagan
  Bansal}, \bibinfo{person}{Cheng Tan}, \bibinfo{person}{Adam Fourney},
  \bibinfo{person}{Victor Dibia}, \bibinfo{person}{Jingya Chen},
  \bibinfo{person}{Jack Gerrits}, \bibinfo{person}{Tyler Payne},
  \bibinfo{person}{Matheus~Kunzler Maldaner}, \bibinfo{person}{Madeleine
  Grunde-McLaughlin}, \bibinfo{person}{Eric Zhu}, \bibinfo{person}{Griffin
  Bassman}, \bibinfo{person}{Jacob Alber}, \bibinfo{person}{Peter Chang},
  \bibinfo{person}{Ricky Loynd}, \bibinfo{person}{Friederike Niedtner},
  \bibinfo{person}{Ece Kamar}, \bibinfo{person}{Maya Murad},
  \bibinfo{person}{Rafah Hosn}, {and} \bibinfo{person}{Saleema Amershi}.}
  \bibinfo{year}{2025}\natexlab{}.
\newblock \bibinfo{booktitle}{\emph{Magentic-UI: Towards Human-in-the-loop
  Agentic Systems}}.
\newblock \bibinfo{type}{{T}echnical {R}eport}. \bibinfo{institution}{Microsoft
  Research}.
\newblock
\newblock
\shownote{arXiv:2507.22358}.


\bibitem[Puthenputhussery et~al\mbox{.}(2025)]%
        {puthenputhussery2025bert}
\bibfield{author}{\bibinfo{person}{Ajit Puthenputhussery},
  \bibinfo{person}{Changsung Kang}, \bibinfo{person}{Alessandro Magnani},
  \bibinfo{person}{Tian Zhang}, \bibinfo{person}{Hongwei Shang},
  \bibinfo{person}{Nitin Yadav}, \bibinfo{person}{Prijith Chandran},
  \bibinfo{person}{Bhavin Madhani}, \bibinfo{person}{Yuan-Tai Fu},
  \bibinfo{person}{He Wang}, \bibinfo{person}{Zbigniew Gasiorek},
  \bibinfo{person}{Salvatore Tornatore}, \bibinfo{person}{Srikanth Dasaka},
  \bibinfo{person}{Vivek Agrawal}, \bibinfo{person}{Michael Bowersox},
  \bibinfo{person}{Cun Mu}, {and} \bibinfo{person}{Ciya Liao}.}
  \bibinfo{year}{2025}\natexlab{}.
\newblock \showarticletitle{Large Scale Deployment of BERT Based Cross Encoder
  Model for Re-Ranking in Walmart Search Engine}. In
  \bibinfo{booktitle}{\emph{Proceedings of the 48th International ACM SIGIR
  Conference on Research and Development in Information Retrieval}}.
  \bibinfo{pages}{4365--4369}.
\newblock
\urldef\tempurl%
\url{https://doi.org/10.1145/3726302.3731965}
\showDOI{\tempurl}


\bibitem[Qian et~al\mbox{.}(2025)]%
        {agentmarketarena2025}
\bibfield{author}{\bibinfo{person}{Lingfei Qian}, \bibinfo{person}{Xueqing
  Peng}, \bibinfo{person}{Yan Wang}, \bibinfo{person}{Vincent~Jim Zhang},
  \bibinfo{person}{Huan He}, \bibinfo{person}{Hanley Smith},
  \bibinfo{person}{Yi Han}, \bibinfo{person}{Yueru He},
  \bibinfo{person}{Haohang Li}, \bibinfo{person}{Yupeng Cao},
  \bibinfo{person}{Yangyang Yu}, \bibinfo{person}{Alejandro Lopez-Lira},
  \bibinfo{person}{Peng Lu}, \bibinfo{person}{Jian-Yun Nie},
  \bibinfo{person}{Guojun Xiong}, \bibinfo{person}{Jimin Huang}, {and}
  \bibinfo{person}{Sophia Ananiadou}.} \bibinfo{year}{2025}\natexlab{}.
\newblock \showarticletitle{When Agents Trade: Live Multi-Market Trading
  Benchmark for LLM Agents}.
\newblock \bibinfo{journal}{\emph{arXiv preprint arXiv:2510.11695}}
  (\bibinfo{year}{2025}).
\newblock


\bibitem[Shinn et~al\mbox{.}(2023)]%
        {reflexion2023}
\bibfield{author}{\bibinfo{person}{Noah Shinn}, \bibinfo{person}{Federico
  Cassano}, \bibinfo{person}{Edward Berman}, \bibinfo{person}{Ashwin Gopinath},
  \bibinfo{person}{Karthik Narasimhan}, {and} \bibinfo{person}{Shunyu Yao}.}
  \bibinfo{year}{2023}\natexlab{}.
\newblock \showarticletitle{Reflexion: Language Agents with Verbal
  Reinforcement Learning}. In \bibinfo{booktitle}{\emph{Advances in Neural
  Information Processing Systems (NeurIPS)}}.
\newblock
\newblock
\shownote{arXiv:2303.11366}.


\bibitem[Subrahmanyam(1994)]%
        {subrahmanyam1994}
\bibfield{author}{\bibinfo{person}{Avanidhar Subrahmanyam}.}
  \bibinfo{year}{1994}\natexlab{}.
\newblock \showarticletitle{Circuit Breakers and Market Volatility: A
  Theoretical Perspective}.
\newblock \bibinfo{journal}{\emph{Journal of Finance}} \bibinfo{volume}{49},
  \bibinfo{number}{1} (\bibinfo{year}{1994}), \bibinfo{pages}{237--254}.
\newblock


\bibitem[{U.S. Securities and Exchange Commission}(2010)]%
        {sec15c3-5}
\bibfield{author}{\bibinfo{person}{{U.S. Securities and Exchange Commission}}.}
  \bibinfo{year}{2010}\natexlab{}.
\newblock \bibinfo{title}{Rule 15c3-5: Risk Management Controls for Brokers or
  Dealers with Market Access}.
\newblock \bibinfo{howpublished}{17 CFR 240.15c3-5}.
\newblock


\bibitem[Wang et~al\mbox{.}(2023)]%
        {voyager2023}
\bibfield{author}{\bibinfo{person}{Guanzhi Wang}, \bibinfo{person}{Yuqi Xie},
  \bibinfo{person}{Yunfan Jiang}, \bibinfo{person}{Ajay Mandlekar},
  \bibinfo{person}{Chaowei Xiao}, \bibinfo{person}{Yuke Zhu},
  \bibinfo{person}{Linxi Fan}, {and} \bibinfo{person}{Anima Anandkumar}.}
  \bibinfo{year}{2023}\natexlab{}.
\newblock \showarticletitle{Voyager: An Open-Ended Embodied Agent with Large
  Language Models}.
\newblock \bibinfo{journal}{\emph{arXiv preprint arXiv:2305.16291}}
  (\bibinfo{year}{2023}).
\newblock


\bibitem[Wang et~al\mbox{.}(2025)]%
        {agentspec2025}
\bibfield{author}{\bibinfo{person}{Haoyu Wang}, \bibinfo{person}{Christopher~M.
  Poskitt}, {and} \bibinfo{person}{Jun Sun}.} \bibinfo{year}{2025}\natexlab{}.
\newblock \showarticletitle{AgentSpec: Customizable Runtime Enforcement for
  Safe and Reliable LLM Agents}.
\newblock \bibinfo{journal}{\emph{arXiv preprint arXiv:2503.18666}}
  (\bibinfo{year}{2025}).
\newblock


\bibitem[Xia et~al\mbox{.}(2026)]%
        {agentictrading2026}
\bibfield{author}{\bibinfo{person}{Yihan Xia}, \bibinfo{person}{Panpan You},
  \bibinfo{person}{Taotao Wang}, \bibinfo{person}{Fang Liu},
  \bibinfo{person}{Han Qi}, \bibinfo{person}{Xiaoxiao Wu}, {and}
  \bibinfo{person}{Shengli Zhang}.} \bibinfo{year}{2026}\natexlab{}.
\newblock \showarticletitle{Agentic Trading: When LLM Agents Meet Financial
  Markets}.
\newblock \bibinfo{journal}{\emph{arXiv preprint arXiv:2605.19337}}
  (\bibinfo{year}{2026}).
\newblock


\bibitem[Xiang et~al\mbox{.}(2024)]%
        {guardagent2024}
\bibfield{author}{\bibinfo{person}{Zhen Xiang}, \bibinfo{person}{Linzhi Zheng},
  \bibinfo{person}{Yanjie Li}, \bibinfo{person}{Junyuan Hong},
  \bibinfo{person}{Qinbin Li}, \bibinfo{person}{Han Xie},
  \bibinfo{person}{Jiawei Zhang}, \bibinfo{person}{Zidi Xiong},
  \bibinfo{person}{Chulin Xie}, \bibinfo{person}{Carl Yang},
  \bibinfo{person}{Dawn Song}, {and} \bibinfo{person}{Bo Li}.}
  \bibinfo{year}{2024}\natexlab{}.
\newblock \showarticletitle{GuardAgent: Safeguard LLM Agents by a Guard Agent
  via Knowledge-Enabled Reasoning}.
\newblock \bibinfo{journal}{\emph{arXiv preprint arXiv:2406.09187}}
  (\bibinfo{year}{2024}).
\newblock


\bibitem[Xiao et~al\mbox{.}(2024)]%
        {tradingagents2024}
\bibfield{author}{\bibinfo{person}{Yijia Xiao}, \bibinfo{person}{Edward Sun},
  \bibinfo{person}{Di Luo}, {and} \bibinfo{person}{Wei Wang}.}
  \bibinfo{year}{2024}\natexlab{}.
\newblock \showarticletitle{TradingAgents: Multi-Agents LLM Financial Trading
  Framework}.
\newblock \bibinfo{journal}{\emph{arXiv preprint arXiv:2412.20138}}
  (\bibinfo{year}{2024}).
\newblock


\bibitem[Yang et~al\mbox{.}(2024)]%
        {finrobot2024}
\bibfield{author}{\bibinfo{person}{Hongyang Yang}, \bibinfo{person}{Boyu
  Zhang}, \bibinfo{person}{Neng Wang}, \bibinfo{person}{Cheng Guo},
  \bibinfo{person}{Xiaoli Zhang}, \bibinfo{person}{Likun Lin},
  \bibinfo{person}{Junlin Wang}, \bibinfo{person}{Tianyu Zhou},
  \bibinfo{person}{Mao Guan}, \bibinfo{person}{Runjia Zhang}, {and}
  \bibinfo{person}{Christina~Dan Wang}.} \bibinfo{year}{2024}\natexlab{}.
\newblock \showarticletitle{FinRobot: An Open-Source AI Agent Platform for
  Financial Applications using Large Language Models}.
\newblock \bibinfo{journal}{\emph{arXiv preprint arXiv:2405.14767}}
  (\bibinfo{year}{2024}).
\newblock


\bibitem[Yu et~al\mbox{.}(2025a)]%
        {shang2025distill}
\bibfield{author}{\bibinfo{person}{Haofei Yu}, \bibinfo{person}{Fenghai Li},
  {and} \bibinfo{person}{Jiaxuan You}.} \bibinfo{year}{2025}\natexlab{a}.
\newblock \showarticletitle{Knowledge Distillation for Enhancing Walmart
  E-commerce Search Relevance Using Large Language Models}. In
  \bibinfo{booktitle}{\emph{Companion Proceedings of the ACM Web Conference
  2025 (WWW Companion)}}. \bibinfo{pages}{449--457}.
\newblock
\newblock
\shownote{arXiv:2505.07105}.


\bibitem[Yu et~al\mbox{.}(2025b)]%
        {livetradebench2025}
\bibfield{author}{\bibinfo{person}{Haofei Yu}, \bibinfo{person}{Fenghai Li},
  {and} \bibinfo{person}{Jiaxuan You}.} \bibinfo{year}{2025}\natexlab{b}.
\newblock \showarticletitle{LiveTradeBench: Seeking Real-World Alpha with Large
  Language Models}.
\newblock \bibinfo{journal}{\emph{arXiv preprint arXiv:2511.03628}}
  (\bibinfo{year}{2025}).
\newblock


\bibitem[Yu et~al\mbox{.}(2023)]%
        {finmem2024}
\bibfield{author}{\bibinfo{person}{Yangyang Yu}, \bibinfo{person}{Haohang Li},
  \bibinfo{person}{Zhi Chen}, \bibinfo{person}{Yuechen Jiang},
  \bibinfo{person}{Yang Li}, \bibinfo{person}{Denghui Zhang},
  \bibinfo{person}{Rong Liu}, \bibinfo{person}{Jordan~W. Suchow}, {and}
  \bibinfo{person}{Khaldoun Khashanah}.} \bibinfo{year}{2023}\natexlab{}.
\newblock \showarticletitle{FinMem: A Performance-Enhanced LLM Trading Agent
  with Layered Memory and Character Design}.
\newblock \bibinfo{journal}{\emph{arXiv preprint arXiv:2311.13743}}
  (\bibinfo{year}{2023}).
\newblock


\bibitem[Zhang et~al\mbox{.}(2024a)]%
        {stockagent2024}
\bibfield{author}{\bibinfo{person}{Chong Zhang}, \bibinfo{person}{Xinyi Liu},
  \bibinfo{person}{Zhongmou Zhang}, \bibinfo{person}{Mingyu Jin},
  \bibinfo{person}{Lingyao Li}, \bibinfo{person}{Zhenting Wang},
  \bibinfo{person}{Wenyue Hua}, \bibinfo{person}{Dong Shu},
  \bibinfo{person}{Suiyuan Zhu}, \bibinfo{person}{Xiaobo Jin},
  \bibinfo{person}{Sujian Li}, \bibinfo{person}{Mengnan Du}, {and}
  \bibinfo{person}{Yongfeng Zhang}.} \bibinfo{year}{2024}\natexlab{a}.
\newblock \showarticletitle{When AI Meets Finance (StockAgent): Large Language
  Model-based Stock Trading in Simulated Real-world Environments}.
\newblock \bibinfo{journal}{\emph{arXiv preprint arXiv:2407.18957}}
  (\bibinfo{year}{2024}).
\newblock


\bibitem[Zhang et~al\mbox{.}(2024b)]%
        {finagent2024}
\bibfield{author}{\bibinfo{person}{Wentao Zhang}, \bibinfo{person}{Lingxuan
  Zhao}, \bibinfo{person}{Haochong Xia}, \bibinfo{person}{Shuo Sun},
  \bibinfo{person}{Jiaze Sun}, \bibinfo{person}{Molei Qin},
  \bibinfo{person}{Xinyi Li}, \bibinfo{person}{Yuqing Zhao},
  \bibinfo{person}{Yilei Zhao}, \bibinfo{person}{Xinyu Cai},
  \bibinfo{person}{Longtao Zheng}, \bibinfo{person}{Xinrun Wang}, {and}
  \bibinfo{person}{Bo An}.} \bibinfo{year}{2024}\natexlab{b}.
\newblock \showarticletitle{A Multimodal Foundation Agent for Financial
  Trading: Tool-Augmented, Diversified, and Generalist}. In
  \bibinfo{booktitle}{\emph{Proceedings of the 30th ACM SIGKDD Conference on
  Knowledge Discovery and Data Mining (KDD)}}.
\newblock
\newblock
\shownote{arXiv:2402.18485}.


\end{thebibliography}

\appendix
\section{Full Incident Catalogue}
\label{app:incidents}

Table~\ref{tab:incidents-full} lists all twenty catalogued incidents and
defects, of which Table~\ref{tab:incidents} in the main text shows ten.
``Adversarial review'' means caught before merge by the independent
reviewer described in \S\ref{sec:builder}; ``operator'' or ``code''
followed by ``production'' means the incident surfaced during live
operation.

\begin{table*}[p]
\caption{All twenty catalogued incidents, in the order they were recorded.}
\label{tab:incidents-full}
\footnotesize
\begin{tabular}{@{}p{0.34\textwidth}p{0.19\textwidth}p{0.41\textwidth}@{}}
\toprule
\textbf{Incident} & \textbf{Caught by} & \textbf{Design rule derived} \\
\midrule
Degenerate action amount crashes strategy loading & Adversarial review & Domain-typed parse boundary, never a bare stdlib exception \\
Condition-evaluator recursion DoS / non-finite literals & Adversarial review & Depth cap + finite-number constraint on any expression language, even agent-authored \\
TOCTOU double-execute race on order claim & Adversarial review & Atomic DB-level claim on every irreversible-action gate \\
Post-claim execution error misreported as "queued" & Adversarial review & Split try/except; every terminal state truthfully recorded \\
Stale-revision silently reverts an approved change & Adversarial review (reproduced pre-merge) & Byte-exact re-check of live file state at approval time, not the proposal's snapshot \\
Draft strategies never monitored, no signal & Operator, production & Persistent, unmissable badge on any state that disables monitoring \\
Advertised-but-unregistered \texttt{session\_search} tool & Adversarial review & Test prompt affordance claims against the actual tool registry \\
Corrective-turn budget silently doubling & Adversarial review & Single enforcement point for a cap, covering retry paths \\
NUL-byte chat message $\to$ persistent 500 & Adversarial review & Strip/reject bytes that collide with internal escaping schemes, from untrusted input including model output \\
Authorization/status freeze on reflection revisions & Code gate (by design) & Structurally exclude "scope of autonomy" fields from what self-revision can write \\
Gmail app-password NBSP corruption & Operator, production & Normalize all whitespace variants on any pasted third-party credential \\
Submitted-vs-filled label drift (partial fix) & Adversarial review, fix was incomplete & Bucket by the event that changed the world (fill), not the event that requested it \\
Options-during-closed-market: id-less 200 payload journaled as submitted & Operator, production (autonomous regime) & Validate response \emph{content} (not just shape) before journaling success; gate scheduled actions on known-good preconditions \\
Options MCP dependency outage (\texttt{uvx} not found), 2.5h retry storm & Code (contained), production & Preflight-check or pin external CLI dependencies; alert distinctly from routine per-order errors \\
\texttt{McpOptionsBackend} subprocess/thread leaks (2 rounds) & Adversarial review (second round) & Owned per-attempt lifecycle object, never shared mutable state across spawn attempts \\
Non-finite \texttt{previous\_close} from quote feed & Code gate, production & Strict validation at ingestion; loud isolated failure beats silent \texttt{NaN} comparison \\
"Analysis unparseable" fails open to human queue & Agent self-check $\to$ code gate default & Never guess or default-execute on unparseable agent output; fail closed to human \\
\texttt{.env} bricked by unvalidated settings save & Adversarial review & Validate the fully assembled state before persisting any part of it \\
Non-finite money values permanently brick shadow ledger & Adversarial review & Validate at every write boundary a value will later be read through \\
\texttt{record\_fill} silently corrupts \texttt{avg\_cost} cross-position & Adversarial review & Guard corrections against intervening state changes on the same key \\
\bottomrule
\end{tabular}
\end{table*}

\section{Artifacts from the Deployment}
\label{app:artifacts}

\paragraph{An approval card, field by field.} Table~\ref{tab:card}
lists what the pending-review page renders for the two kinds of card,
with the values from the repository's demo data.

\begin{table*}[t]
\caption{Fields of an approval card (demo data). The diff is regenerated
against the current file at render time; the order card's live price and
deviation are recomputed on every view.}
\label{tab:card}
\small
\begin{tabular}{@{}p{0.22\textwidth}p{0.74\textwidth}@{}}
\toprule
\textbf{Field} & \textbf{Rendered value} \\
\midrule
\multicolumn{2}{@{}l}{\emph{Strategy revision card}} \\
Header & revise strategy $\cdot$ NVDA \#2 $\cdot$ \texttt{reflection} \\
Rationale & ``NVDA has re-rated: 140 is now 23\% below spot and would only fire after a crash we could not act on. Tightening the hard stop to 150 keeps the thesis intact while capping drawdown near 18\%.'' \\
Diff, current vs.\ proposed & \texttt{version: 3} $\to$ \texttt{4}; \texttt{stop} rule \texttt{price < 140} $\to$ \texttt{price < 150}; unchanged lines collapsed (``\ldots\ 5 unchanged lines'') \\
Frozen fields shown & \texttt{status: active}, \texttt{authorization: confirm} (unchanged by construction) \\
Controls & Approve; Reject with optional reason \\
\midrule
\multicolumn{2}{@{}l}{\emph{Order card}} \\
Header & buy \$2000 $\cdot$ AMD \#1 \\
Source rule & \texttt{amd-swing/buy-dip}, triggered on \texttt{price < 155} \\
Trigger-time price & \$154.62 \\
Live price & \$164.15 ($-$1.9\% today) \\
Deviation since trigger & $+$6.2\% \\
Estimated shares & $\approx$12.18 at current price \\
Market state & ``Market open: fills near current price'' (or ``market closed: fills at next open'') \\
Controls & Approve; Reject with optional reason \\
\bottomrule
\end{tabular}
\end{table*}

\paragraph{A lesson entry.} One entry from the paper account's
\texttt{lessons} memory layer, written by the agent after a sequence of
fills landed on the wrong side of their triggers (verbatim):

\begin{quote}\small
Confirm-mode approval lag (\textasciitilde20--120 min) on high-ATR names
lands fills on the wrong side of the trigger (MU 9/21 TP @ \$1,038 vs
$>$1050; 10/7 dip-buy @ \$1,034 vs $<$1030). Read triggers as
\textasciitilde$\pm$1\% zones (\textasciitilde0.3 ATR), and don't treat
fills inside that zone as a rule flaw.
\end{quote}

\noindent The entry records the same human-latency effect quantified in
\S\ref{sec:evidence}, observed by the agent from its own fills.

\paragraph{A reflection report excerpt.} The opening of the paper
account's nightly report for 2026-10-08 (verbatim, truncated):

\begin{quote}\small
\textbf{Day summary (2026-10-08, paper):} No rules fired and nothing is
pending. Chip stocks fell together. News reports blamed higher bond
yields and oil prices, plus Samsung's Q3 operating profit, which jumped
but still came in under very high expectations. MU fell $-$4.79\% to
close at \$1,035.84, AMD $-$3.90\% to \$620.68, TSM $-$3.00\% and NVDA
$-$2.94\%. [\ldots] Every position still shows an open gain. Caveat: the
position values the portfolio tool reported look the same as the
session-start snapshot. They may not include today's close yet, so treat
the equity figure as approximate.

\textbf{Per-strategy check:}
\begin{itemize}
\item \textbf{mu-oversold-swing:} Today wiped out most of 10/7's rebound.
  The 10.09 shares bought at \$1,032.67 average are now about breakeven,
  at \textasciitilde10\% weight against a 10\% target. No rule is close
  to firing: dip-buy-1 is blocked by its weight cap; dip-buy-2
  ($<$\$970) is 6.4\% below; the stop ($<$\$900) is 13\% below. [\ldots]
\end{itemize}
\end{quote}

\noindent The report then checks each remaining strategy against the
day's prices and triggers, and closes with memory updates and, when
warranted, a revision proposal. Every report in the window is published
on the public journal page.

\end{document}